\documentclass[10pt,twocolumn,letterpaper]{article}

\usepackage{iccv}              

\usepackage{colortbl}

\definecolor{iccvblue}{rgb}{0.21,0.49,0.74}
\usepackage[pagebackref,breaklinks,colorlinks,allcolors=iccvblue]{hyperref}

\def\paperID{218} 
\def\confName{ICCV}
\def\confYear{2025}

\title{CA-I2P: Channel-Adaptive Registration Network with Global Optimal Selection}

\author{
Zhixin Cheng$^{1}$ \quad Jiacheng Deng$^{1}$ \quad Xinjun Li$^{1}$ \quad Xiaotian Yin$^{2}$ \\
Bohao Liao$^{1}$ \quad Baoqun Yin$^{1}$ \quad Wenfei Yang$^{1}$ \quad Tianzhu Zhang$^{1}$\thanks{Corresponding author.} \\
$^1$School of Information Science and Technology, University of Science and Technology of China \\
$^2$Institute of Advanced Technology, University of Science and Technology of China \\
{\tt\small \{chengzhixin, dengjc, lxj3017, xiaotianyin, liaobh\}@mail.ustc.edu.cn} \\
{\tt\small \{bqyin, yangwf, tzzhang\}@ustc.edu.cn}
}

\begin{document}
\maketitle
\begin{abstract}
Detection-free methods typically follow a coarse-to-fine pipeline, extracting image and point cloud features for patch-level matching and refining dense pixel-to-point correspondences. However, differences in feature channel attention between images and point clouds may lead to degraded matching results, ultimately impairing registration accuracy. Furthermore, similar structures in the scene could lead to redundant correspondences in cross-modal matching. To address these issues, we propose Channel Adaptive Adjustment Module (CAA) and Global Optimal Selection Module (GOS). CAA enhances intra-modal features and suppresses cross-modal sensitivity, while GOS replaces local selection with global optimization. Experiments on RGB-D Scenes V2 and 7-Scenes demonstrate the superiority of our method, achieving state-of-the-art performance in image-to-point cloud registration.
\end{abstract}    
\section{Introduction}
\label{sec:intro}

\begin{figure}[ht]
    \centering
    \includegraphics[width=1\linewidth]{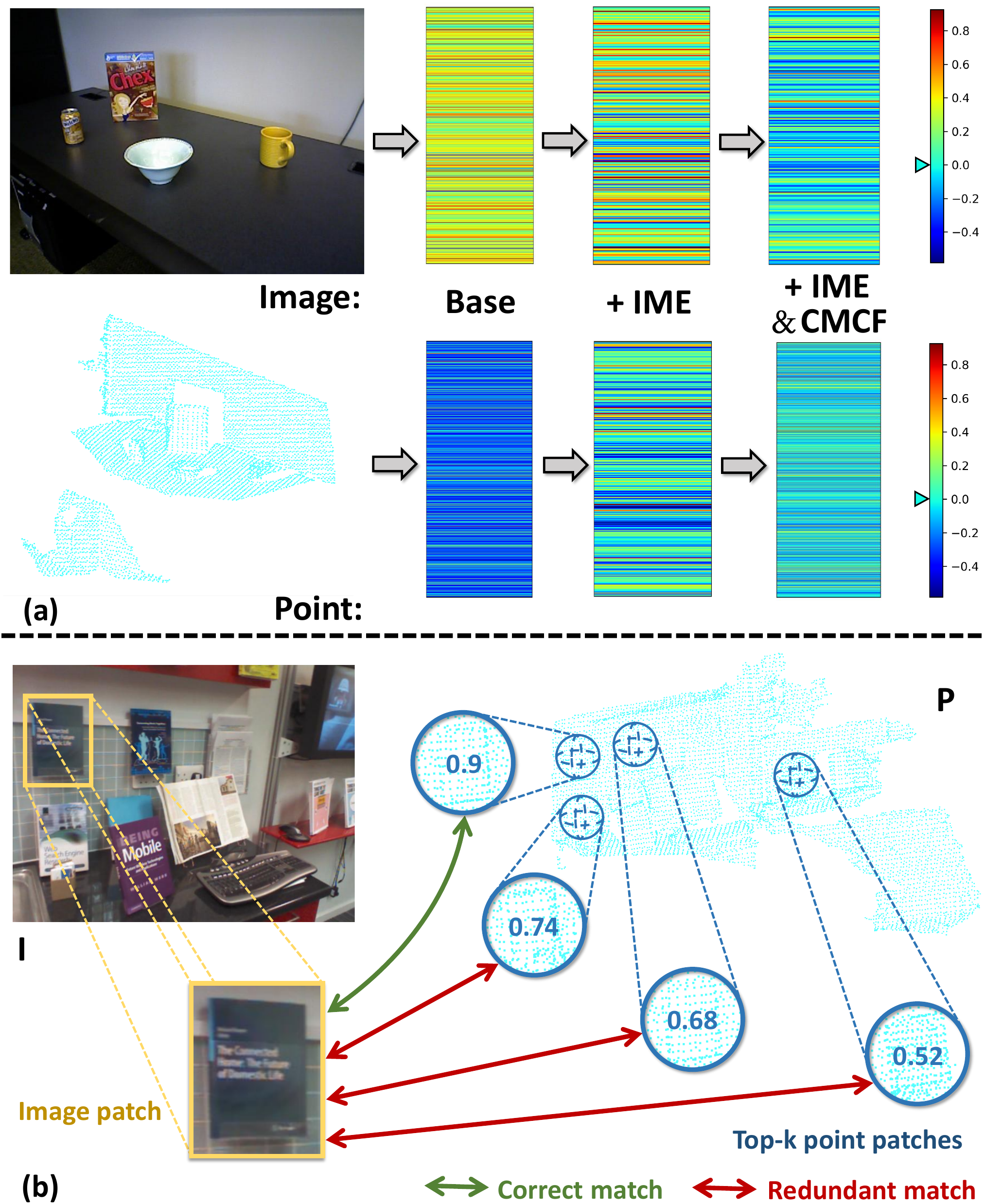} 
    \caption{(a) Visualization of feature variations across channels. Intra-Modal Enhancement stage (IME) and Cross-Modal Channel Filtering stage (CMCF) align feature distributions, reducing modal differences and enhancing matching accuracy and robustness, as reflected in the colors becoming more uniform.
    (b) Visualization of patch-level selection process: top-k causes redundant matches, while we minimizes many-to-one errors through targeted design.}
    \label{fig:1}
    \vspace{-8pt}
\end{figure}

Image-to-point cloud registration (I2P) estimates the rigid transformation from the point cloud reference frame to the camera reference frame by matching pixels and points across modalities and computing the rotation and translation matrices. It is crucial for various computer vision tasks, such as 3D reconstruction \cite{3dreconstruction,deng2024unsupervised,dengseornet}, SLAM \cite{slam,wanghao}, and visual localization \cite{denghierarchical,deng2025quantity}. However, images are 2D data represented as regular, dense grids, while point clouds are unordered, sparse, and irregular 3D data. The domain gap between image features from a 2D encoder and point cloud features from a 3D encoder often manifests in feature channels, posing a significant challenge.

To address the challenges in image-to-point cloud registration, methods can be broadly categorized into two types. The first type is the detect-then-match method \cite{2d3dmatchnet, p2, corri2p}, which involves detecting 2D and 3D keypoints separately in the image and point cloud, and then matching them based on their associated descriptors. However, this approach encounters two main issues: firstly, it is difficult to detect consistent keypoints due to the differing characteristics of 2D and 3D data; and then, matching pixels and points is complicated by the use of inconsistent 2D and 3D descriptors. As a result, detection-free methods \cite{cofii2p, matr2d3d} have become more prevalent.
 2D3D-MATR \cite{matr2d3d} adopts a coarse-to-fine approach, initially establishing patch-level correspondences between image and point cloud features, refining them to pixel-to-point matches, and finally computing the transformation using PnP-RANSAC \cite{pnp,ransac}. However, differences in the imaging ranges of LiDAR and cameras lead to channel-level discrepancies between the two modalities, which may affect the accuracy of cross-modal matching.

By discussing detection-free methods, we highlight two key challenges that need to be addressed for accurate and reliable image-to-point cloud matching. \textbf{Firstly, enhancing and filtering features at the channel level for both data is essential for improving the representation of matching regions.} Previous methods have improved registration accuracy by constructing dense matching within patches. Still, mismatches can occur due to differences in feature attention across channels, leading to misaligned receptive fields and occlusions \cite{zhang2022revisiting}.
Figure~\ref{fig:1}(a) illustrates the differences in channels between the two modalities through colors, which may result in incorrect matches. After completing our module, modal consistency gradually improves.
\textbf{Secondly, a global optimization approach is needed to improve the selection process and reduce redundant correspondences in cross-modal matching.} 
Scenes often have similar structures, and the previous top-k selection could lead to multiple similar structures being incorrectly matched to the same cross-modal object. Therefore, we need a global optimization strategy to replace the local pixel-by-point selection method \cite{ot,sinkhorn}.
As seen in Figure \ref{fig:1}(b), similarity values (numbers on point cloud patches) show that top-k selection can lead to redundant matches (red lines), potentially affecting correct matches (green lines).

To address the challenges mentioned, we propose the Channel-Adaptive Registration Network with Global Optimal Selection (CA-I2P), introducing two innovative modules: the Channel Adaptive Adjustment Module (CAA) and the Global Optimal Selection Module (GOS). In the CAA, we present the Intra-Modal Enhancement Stage (IME) and Cross-Modal Channel Filtering Stage (CMCF) to enhance and filter the image and point cloud features at the channel level. In the IME, for image features, we use rotation and residual transformations to establish inter-channel dependencies \cite{senet,cbam,triplet}, while for point cloud features, we apply self-attention \cite{transformer} mechanisms to enhance the channel dimension. CMCF identifies and masks incompatible channels through the covariance matrix. Then, we merge the processed features with the original ones, which helps focus on matching regions while maintaining feature independence and reducing domain discrepancies. In GOS, we replace the previous top-k selection method with optimal transport. This global approach effectively reduces multi-to-one matching errors, replacing the local optimization with a global strategy to improve the accuracy and robustness of image-to-point cloud registration.

In summary, our work can be outlined as follows:

\begin{enumerate}
    \item We propose a novel Channel-Adaptive Registration Network with Global Optimal Selection that achieves excellent accuracy and strong generalization in image-to-point cloud registration tasks.
    \item We design the Channel Adaptive Adjustment Module, which enhances and filters image and point cloud features at the channel level, both within and across modalities. This improvement aids in better representation for matching regions and reduces domain discrepancies. Additionally, we introduce the Global Optimal Selection Module to minimize multi-to-one errors in matching between modalities, thereby enhancing global consistency.
    \item Extensive experiments and ablation studies on two benchmarks, RGB-D Scenes V2 and 7-Scenes, demonstrate the superiority of our method, establishing it as a state-of-the-art approach in I2P registration.
\end{enumerate}
\section{Related Works}
In this section, we briefly overview related works on I2P registration, including stereo image registration, point cloud registration, and inter-modality registration. 

\subsection{Stereo Image Registration}

Traditional stereo image registration has long relied on detector-based methods, using handcrafted keypoint detection and description for feature matching. Techniques like SIFT \cite{sift} and ORB \cite{orb} have been key for 2D image matching. With deep learning, neural network-based detectors and descriptors, such as SuperGlue \cite{superglue}, have improved feature matching using transformers \cite{transformer}. However, these methods struggle with repeatable keypoints in non-prominent regions, limiting robustness. To address this, detector-free methods like LoFTR \cite{loftr} and Efficient LoFTR \cite{efficientloftr} adopt a coarse-to-fine pipeline with global receptive fields, enabling more accurate dense matching.

\subsection{Point Cloud Registration}

Point cloud registration has evolved from handcrafted descriptors like PPF \cite{ppf} and FPFH \cite{fpfh} to deep learning-based approaches. CoFiNet \cite{cofinet} pioneered the coarse-to-fine process, driving detector-free registration methods. Recent research focuses on replacing traditional estimators like RANSAC \cite{ransac} with deep learning-based alternatives for faster and more accurate registration. GeoTransformer \cite{geotransformer} enhances inlier rates using transformers to capture global structural information and introduces a local-to-global registration framework (LGR) that enables RANSAC-free registration.

\subsection{Inter-Modality Registration}
Inter-modal registration presents more significant challenges than intra-modal registration due to differences in data domains. Traditional methods typically follow a detect then match approach. For instance, 2D3DMatch-Net \cite{2d3dmatchnet} extracts key points using SIFT \cite{sift}, and ISS \cite{iss} builds patches and employs CNNs and PointNet \cite{pointnet} for feature extraction and matching. P2-Net \cite{p2} enhances efficiency by utilizing a joint learning framework that detects key points and matches them in a single pass. However, the precision of keypoint detection \cite{keypoint,keypoint2} is often compromised in cross-modal scenarios, which limits overall performance and has led to the development of detector-free methods \cite{freereg}.
2D3D-MATR \cite{matr2d3d} implements a coarse-to-fine strategy, using transformers for initial coarse matching and refinement using PnP and RANSAC \cite{pnp,ransac} for transformation. This method eliminates the need for keypoint detection and facilitates better alignment of descriptors. Our approach, CA-I2P, is also detector-free and addresses issues related to feature bias and multi-to-one matching through global smoothing, establishing a new state-of-the-art image-to-point cloud registration.
\section{Method}

\begin{figure*}[ht]
    \centering
    \includegraphics[width=\textwidth]{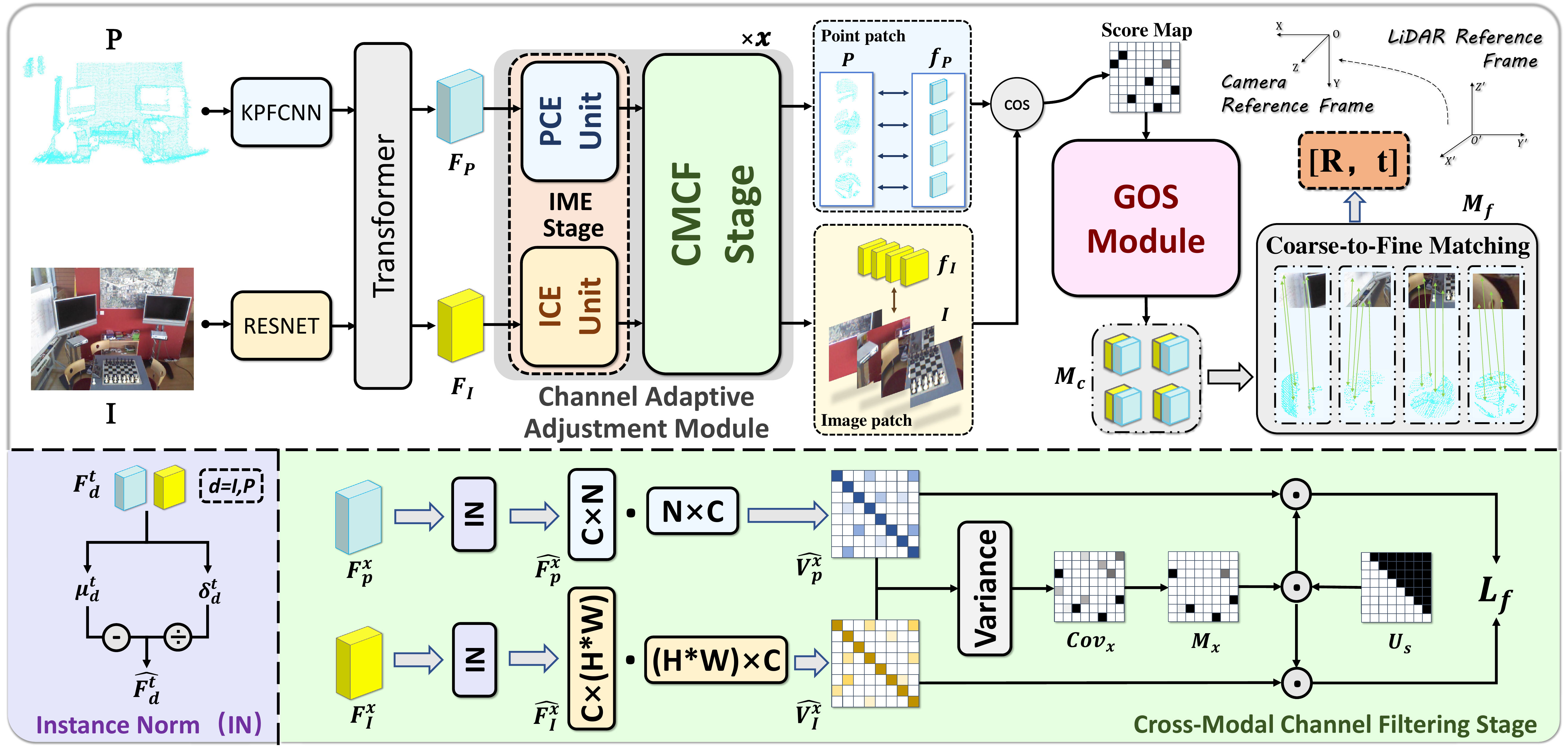} 
    \caption{Overall pipeline of CA-I2P. We use a feature extraction backbone to obtain features from images and point clouds. The image and point cloud features are processed through the Channel Adaptive Adjustment Module (CAA), which includes the Intra-Modal Enhancement stage (IME) and Cross-Modal Channel Filtering stage (CMCF) to optimize features at the channel level. After computing the cosine similarity to obtain the score map, the Global Optimal Selection Module (GOS) precisely achieves coarse-level matching \(M_c\) and refines fine-level matches \(M_f\). Finally, PnP+RANSAC is used to regress the rigid transformation.
}
    \label{fig:2}
    \vspace{-10pt}
\end{figure*}

\subsection{Overall}
Given a pair of images and point clouds in the same scene, denoted as
\(\mathbf{I} \in \mathbb{R}^{h \times w \times 3}\) and \(\mathbf{P} \in \mathbb{R}^{n \times 3}\),
where \(h\) and \(w\) represent the image height and width, and \(n\) is the number of points.
The goal of image-to-point cloud registration is to estimate a rigid transformation $[\mathbf{R}, \mathbf{t}]$, where the rotation matrix $\mathbf{R} \in \mathit{SO}(3)$
 and the translation vector $\mathbf{t} \in \mathbb{R}^3$.


Our CA-I2P method (as shown in Figure \ref{fig:2}) follows a detector-free matching paradigm with channel-wise enhancement and filtering. The Channel Adaptive Adjustment Module (CAA) refines image and point cloud features at the channel level, reducing bias and improving cross-modal matching. After filtering and similarity computation, the score map is globally optimized by the Global Optimal Selection Module (GOS) to obtain coarse matches $\mathbf{M}_c$, from which dense matches $\mathbf{M}_f$ are extracted. Finally, PnP + RANSAC estimates the rigid transformation.

\subsection{Channel Adaptive Adjustment Module}
We use FPN-equipped \cite{fpn} ResNet \cite{resnet} and KPFCNN \cite{kpfcnn} to extract image and point cloud features separately. We augment the 2D and 3D features with their positional information before the attention layer.
\begin{align}
\hat{F}^{\mathcal{I}}_{\text{pos}} &= \hat{F}^{\mathcal{I}} + \phi(\hat{Q}), \notag \\
\hat{F}^{\mathcal{P}}_{\text{pos}} &= \hat{F}^{\mathcal{P}} + \phi(\hat{P}), 
\end{align}

The Fourier embedding function \(\phi(x)\) \cite{fourier} encodes positional information by transforming it into a sequence of sine and cosine terms :
\begin{equation}
\begin{aligned}
\phi(x) = \big[ & x, \sin(2^0 x), \cos(2^0 x), \ldots, \\
                & \sin(2^{L-1} x), \cos(2^{L-1} x) \big],
\end{aligned}
\end{equation}

where \(L\) is the length of the embedding. This transformation incorporates spatial positioning into the features. To facilitate further computations, the first two spatial dimensions of the 2D features are flattened, making the augmented features \(\hat{F}^{\mathcal{I}}_{\text{pos}}\) and \(\hat{F}^{\mathcal{P}}_{\text{pos}}\) ready for subsequent processing.
After positional encoding \cite{fourier} and transformer \cite{transformer}  enhancement, the features are downsampled to \( F_I \in \mathbb{R}^{H \times W \times C} \) and \( F_P \in \mathbb{R}^{N \times C} \). 


For \( F_I \) and \( F_P \), we aim to optimize them along the channel dimension to mitigate inconsistencies in attention regions caused by differences in image and point cloud features. Such inconsistencies may lead to mismatches in non-overlapping regions, errors in receptive fields, and ambiguities in occlusion-based patch-level matching. By reducing the domain gap between image and point cloud features, our approach improves matching success rates while minimizing interference from non-matching region features.

\subsubsection{Intra-Modal Enhancement Stage}
\begin{figure*}[ht]
    \centering
    \includegraphics[width=\textwidth]{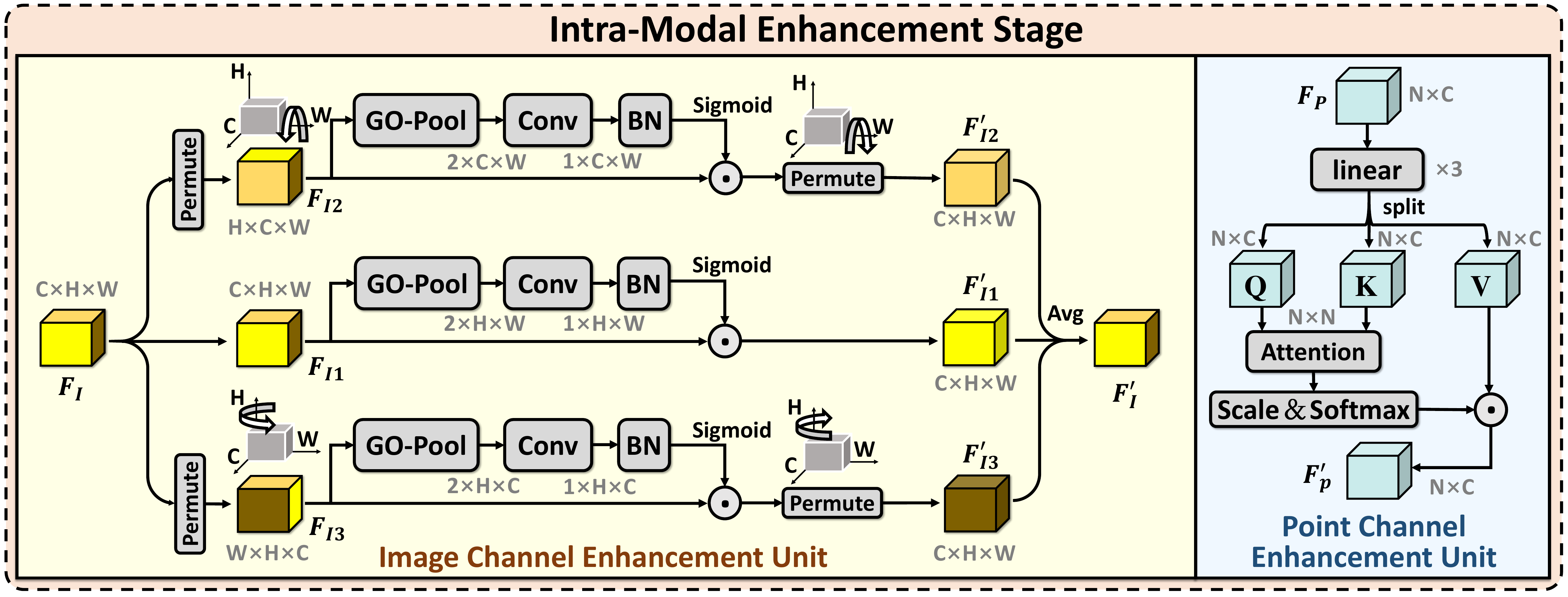} 
    \caption{ The architecture of Intra-Modal Enhancement Stage. It includes the Image Channel Enhancement Unit (ICE) and the Point Channel Enhancement Unit (PCE), which are designed differently based on the characteristics of different features. In ICE, we model the dependencies between the dimensions \((H, W)\), \((C, H)\), and \((C, W)\) to better utilize the input image features. In PCE, we enhance the point cloud features using an self-attention mechanism.
}
    \label{fig:3}
    \vspace{-10pt}
\end{figure*}

We begin by enhancing the features along the channel dimension for each modality. To preserve feature independence, we combine the enhanced features with the original ones. Given the differences between image and point cloud features, we designed the Image Channel Enhancement Unit (ICE) and the Point Cloud Channel Enhancement Unit (PCE), tailoring them to the specific characteristics of each modality. The architecture can be seen in Figure \ref{fig:3}.

\textbf{Image Channel Enhancement Unit.}
We first adjust the image feature channel order to \( F_I \in \mathbb{R}^{C \times H \times W} \). Three parallel branches are designed for the image features: two capture cross-dimensional interactions between the channel dimension \( C \) and spatial dimensions \( H \) or \( W \), and the third establishes spatial attention \cite{cbam}. The final output is the average of the three branches. This approach optimizes attention computation by capturing the relationships between the spatial and channel dimensions.

In ICE, we model the dependencies between the dimensions \( (H, W) \), \( (C, H) \), and \( (C, W) \) to utilize input features better, improving feature representation, cross-modal matching accuracy, and robustness in the registration task.

The GO-Pool layer combines max pooling \cite{maxpool} and average pooling \cite{avgpool} to reduce the first dimension of the tensor to two, preserving feature richness while improving computational efficiency. Specifically, the GO-Pool operation is defined as:
\begin{equation}
    \text{GO-Pool}(F_{I y}) = 
    \begin{aligned}
        &\left[ \text{MaxPool}_{0d}(F_{I y}), \right. \\
        &\quad \left. \text{AvgPool}_{0d}(F_{I y}) \right],
    \end{aligned}
\end{equation}
where \( F_{I y} \) represents the image features of the \( I \)-th feature channel and \( y \)-th branch, and \( 0d \) denotes pooling along the first dimension. \( y \in \{1, 2, 3\} \) represents the different branches.

Taking \( F_{I 2} \) as an example, we establish an interaction between the height and channel dimensions. We rotate \( F_I \) counterclockwise by 90° along the \( W \)-axis. The rotated tensor \( F_{I 2} \) is passed through the GO-Pool layer, reducing it to shape \( 2 \times C \times W \). A convolution layer with kernel size \( k \times k \), followed by a batch normalization (BN) layer, normalizes the output to shape \( 1 \times C \times W \). The tensor is then passed through a sigmoid activation layer \( \sigma \) to obtain attention weights, which are applied to \( F_{I 2} \). Finally, the output is rotated 90° clockwise along the \( W \)-axis to match the original shape of \( F_I \). Similarly, \( F_{I 3} \) is obtained by rotating \( F_I \) counterclockwise along the \( H \)-axis, and \( F_{I 1} \) is not rotated.

This process is expressed as:
\begin{equation}
F_I' = \frac{1}{3} \sum_{y=1}^3 \left( F_{I y} \, \sigma(\text{Conv}(\text{GO-Pool}(F_{I y}))) \right),
\end{equation}
where \( \sigma \) is the sigmoid function, and \( \text{Conv} \) is a 2D convolution layer with kernel size \( k \).

\textbf{Point Channel Enhancement Unit.}
For point cloud features \( F_P \in \mathbb{R}^{N \times C} \), which only have two dimensions, we employ a channel self-attention \cite{vit} mechanism to enhance the point cloud features. This is achieved by calculating the inter-channel correlation of the input features and adaptively adjusting the feature representation to suit the matching task better.

Given the input point features \( F_P \in \mathbb{R}^{N \times C} \), we first project them into three feature representations: query (Q), key (K), and value (V) through a linear layer:
\begin{equation}
Q, K, V = W_Q F_{P 1}, W_K F_{P 2}, W_V F_{P 3},
\end{equation}
where \( W_Q, W_K, W_V \in \mathbb{R}^{C \times C} \) are learnable linear projections. Next, we calculate the attention weights based on the inter-channel relationships of the point features. Using this attention matrix, we perform a weighted sum on the values \( V \) to obtain the enhanced features:
\begin{equation}
A = \text{Softmax}\left(\frac{Q K^T S}{\sqrt{C}}\right),
\end{equation}
\begin{equation}
F_P' = A V,
\end{equation}
where \( A \in \mathbb{R}^{N \times N} \) is the attention matrix between points, and \( S \) is a scaling factor used to stabilize the gradients. 

Finally, the enhanced features \( F_P' \) and \( F_I' \) are weighted and fused with the initial features \( F_I \) and \( F_P \) as follows:
\begin{equation}
F_I = \alpha F_I + \beta F_I',
\end{equation}
\begin{equation}
F_P = \lambda F_P + \mu F_P',
\end{equation}
where \( \alpha, \beta, \lambda, \mu \) are learnable parameters. By updating the image and point cloud features, we enhance their expressiveness to better adapt to cross-modal matching tasks.

\subsubsection{Cross-Modal Channel Filtering Stage}
After enhancement by IME, the representational ability of the image and point cloud features has been improved. However, the consistency of cross-modal features becomes the main obstacle to further improving generalization performance. Typically, I2P registration networks use batch normalization (BN) as the default normalization operation. Batch normalization relies on statistics from the training set, making the model sensitive to changes in data distribution. To enhance cross-domain consistency, we use instance normalization (IN), which normalizes each sample individually, making it independent of the training set statistics and improving generalization ability.

For each sample \( F_d^t \), we normalize it to obtain \( \hat{F_d^t} \):
\begin{equation}
\hat{F_d^t} = \frac{F_d^t - \mu_d^t}{\sigma_d^t},
\end{equation}
where \( \mu_d^t \) and \( \sigma_d^t \) are the mean and standard deviation along the channel index \( t \), and \( d \in \{I, P\} \) for image and point cloud features.

We further consider the information stored in the feature covariance, which has not been processed by IN. Our goal is to suppress feature covariance \cite{conv} components that are sensitive to different modality variations through cross-modal channel filtering, thereby learning common representations shared across different modalities. To achieve this, we compute the covariance matrices for image and point cloud features, denoted as \( \hat{\mathbf{V}}_I^x \in \mathbb{R}^{c \times c} \) and \( \hat{\mathbf{V}}_P^x \in \mathbb{R}^{c \times c} \), respectively. These covariance matrices are obtained using the following formulations:
\begin{equation}
\hat{\mathbf{V}}_I^x = \frac{1}{HW} \left( \hat{\mathbf{F}}_I^{x\top} \cdot \hat{\mathbf{F}}_I^x \right),
\end{equation}
\begin{equation}
\hat{\mathbf{V}}_P^x = \frac{1}{N} \left( \hat{\mathbf{F}}_P^{x\top} \cdot \hat{\mathbf{F}}_P^x \right),
\end{equation}
where \( x \) represents the iteration processed by CMCF. We obtain the final variance matrix \( \mathbf{V}_I^x \in \mathbb{R}^{B \times C \times C} \) by gathering \( \hat{\mathbf{V}}_I^x \) from \( B \) samples in a batch, where \( B \) is the batch size. Similarly, \( \mathbf{V}_P^x \) is obtained by gathering the variance matrix of features \( \hat{\mathbf{F}}_P^x \). Then we compute the covariance matrix \( \mathbf{Cov_x} \in \mathbb{R}^{C \times C} \) between \( \mathbf{V}_I^x \) and \( \mathbf{V}_P^x \).
\begin{equation}
\mu_x = \frac{1}{2B} \sum_{b=1}^{B} (V_I^x (b) + V_P^x (b)),
\end{equation}
\begin{equation}
\mathbf{Cov}_x = \frac{1}{2B} \sum_{b=1}^{B} ((V_I^x (b) - \mu_x)^2 + (V_P^x (b) - \mu_x)^2),
\end{equation}
where \( b \) indexes the samples. The element \( \mathbf{Cov}_x (i, j) \) in the covariance matrix measures how sensitive the \( i \)-th and the \( j \)-th channels are across different modalities. More directly, a higher \( \mathbf{Cov}_x (i, j) \) indicates that the feature correspondence between the \( i \)-th channel of the image and the \( j \)-th channel of the point cloud is more sensitive to variations across modalities, and vice versa. Feature channels with significantly large variance values may contain more attention to non-corresponding regions, which is unfavorable for the image-point cloud registration task. These should be suppressed through constraints; therefore, we design \( L_f \) to filter out parts with excessively high variance values.
\begin{equation}
L_f = \frac{1}{X} \sum_{x=1}^{X} \left( \|\hat{\mathbf{V}}_I^x \odot M_x\|_1 + \|\hat{\mathbf{V}}_P^x \odot M_x\|_1 \right),
\end{equation}
Among them, \( M_x \in (0,1)^{c \times c} \) is a selective mask, where \( x \) represents the layer processed by CMCF. 

The covariance matrix \( \mathbf{Cov}_x \) is divided into three groups based on the width of variance. We select the one with the highest variance value for masking \cite{zhang2022revisiting}. Then, we multiply the selected \( M_x \) by a strict upper triangular matrix \( \mathbf{U}_s \), update \( M_x \), and then apply it to \( \hat{\mathbf{V}}_I^x \) and \( \hat{\mathbf{V}}_P^x \). Since the covariance matrix is symmetric, as training progresses, it may cause the network to overfit the statistical characteristics of a specific modality, making it difficult to generalize to cross-modal scenarios. By optimizing only the upper triangular part, we can prevent the model from overly relying on the statistical information of a particular modality, reduce redundant information, enhance feature independence, and improve the robustness of cross-modal matching. Furthermore, to ensure feature completeness and prevent important information from being masked out, we also integrate the filtered features with the original features, obtaining the image and point cloud features \( f_I \) and \( f_P \).

\subsection{Global Optimal Selection Module}
The previous method calculates the similarity between image and point cloud features and then selects top-\( k \) \cite{topk} matching pairs at the patch level. However, this approach may lead to multiple similar structures being matched to the same cross-modal counterpart. Therefore, hierarchical matching should avoid the habitual reliance on seeking the best match for each pixel and point individually and instead shift toward a globally optimal approach.

Through iterative feature augmentation and filtering updates, we obtain the cross-modal feature interaction matrices between image pixels and point cloud points, denoted as \( [I, f_I] \) and \( [P, f_P] \). Subsequently, the redundancy correlation score map \( S \) is computed. During the inference process, based on the nearest neighbor principle in feature space, we construct a pixel-to-point correspondence between set \( I \) and set \( P \). Each pixel \( I_m \) in \( I \) is assigned to its closest point \( P_n \) in \( P \) based on feature space proximity. This selection rule can be expressed as follows:
\begin{equation}
f(I_m) = P_n, \quad n^* = \arg\max_n (S_{m,n}),
\end{equation}
in the absence of point-level supervision, the learned similarity matrix may produce ambiguous activations. To optimize the similarity matrix online, we propose searching for globally high-confidence matches in the image-point cloud. This problem is equivalent to the Optimal Transport (OT) \cite{ot,ot2} problem, where the goal is to find a transport plan \( T^* \) that minimizes the global transportation cost. This problem can be solved using the Sinkhorn algorithm combined with linear programming.

Given the image representation \( F_I \in \mathbb{R}^{(H \times W) \times C} \), we focus on projecting the point cloud features \( F_P \in \mathbb{R}^{N \times C} \) onto the image. We denote this projection as \( T^* \), where the similarity cost matrix is defined as \( (1 - S) \). The optimal transport plan \( T^* \in \mathbb{R}^{(H \times W) \times N} \) is obtained by minimizing the similarity cost with an additional regularization term:
\begin{equation}
T^* = \min_{T \in \mathcal{T}} \operatorname{Tr} \left( T^T (1 - S) \right) - \epsilon H(\mathcal{T}),
\label{eq15}
\end{equation}
where \( H(\mathcal{T}) \) represents the regularization term. 
Among them, \(H(\mathcal{T}) = -\sum_{ij} T_{ij} \log T_{ij} \) is the entropy function, and \( T \) is the search space. According to the methods in references \cite{ot2,ot3}, \( \mathcal{T} \) is constrained as 
\begin{equation}
\mathcal{T} = \{ T \in \mathbb{R}^{N \times N} \mid T \mathbf{1} = \frac{1}{N} \cdot \mathbf{1}, T^T \mathbf{1} = \frac{1}{N} \cdot \mathbf{1} \},
\end{equation}
where \( \mathbf{1} \) represents an all-ones vector.

The problem described in Equation \ref{eq15} can be efficiently solved using the existing GPU-based Sinkhorn iterations. Research shows that approximately 10 iterations are sufficient to achieve satisfactory results. After maximizing the selection, we obtain accurate patch-level image-to-point cloud correspondences \( \mathbf{M}_c \).

\subsection{Model Training \& Post-Processing Details}
Let us examine the loss functions for the coarse and fine-matching networks. Both \( \mathcal{L}_{\text{coarse}} \) and \( \mathcal{L}_{\text{fine}} \) utilize a general circle loss \cite{circleloss,circleloss1}. For a given anchor descriptor \( d_i \), the descriptors of its positive and negative pairs are represented as \( \mathcal{D}_i^P \) and \( \mathcal{D}_i^N \), respectively. The loss function is defined as follows:
\begin{small}
\begin{equation}
\begin{aligned}
\mathcal{L}_i = \frac{1}{\gamma} \log\biggl[1 + \Bigl(&\sum_{d^j \in \mathcal{D}^P_i} e^{\beta^{i,j}_p(d^j_i - \Delta_p)}\Bigr) \\
&\cdot \Bigl(\sum_{d^k \in \mathcal{D}^N_i} e^{\beta^{i,k}_n(\Delta_n - d^k_i)}\Bigr)\biggr],
\end{aligned}
\end{equation}
\end{small}
where \( d_i^j \) is the \( L_2 \) feature distance, \( \beta_p^{i,j} = \gamma \lambda_p^{i,j} (d_i^j - \Delta_p) \), and \( \beta_n^{i,k} = \gamma \lambda_n^{i,k} (\Delta_n - d_i^k) \) are the individual weights for the positive and negative pairs, with \( \lambda_p^{i,j} \) and \( \lambda_n^{i,k} \) as scaling factors. 
Based on the above discussion, the total loss consists of three key components: the constraint between image and point cloud channels \( L_f \), and the matching process losses \( L_{ic} \) and \( L_{if} \), calculated as:
\begin{equation}
L_{\text{total}} = \lambda_1 L_f + \lambda_2 L_{ic} + \lambda_3 L_{if},
\end{equation}
where \( \lambda_i \) are hyperparameters balancing the contribution of different loss terms.
Next, the patch-level image-to-point cloud correspondences \( \mathbf{M}_c \) are obtained. High-resolution features extracted by the feature extractor are introduced for dense matching \( \mathbf{M}_f \). Finally, the PnP-RANSAC \cite{pnp,ransac} algorithm estimates the rigid transformation between the image and point cloud.

\subsection{Discussion}
\begin{figure}[ht]
  \vspace{-10pt}
  \centering

  \includegraphics[width=0.8\linewidth]{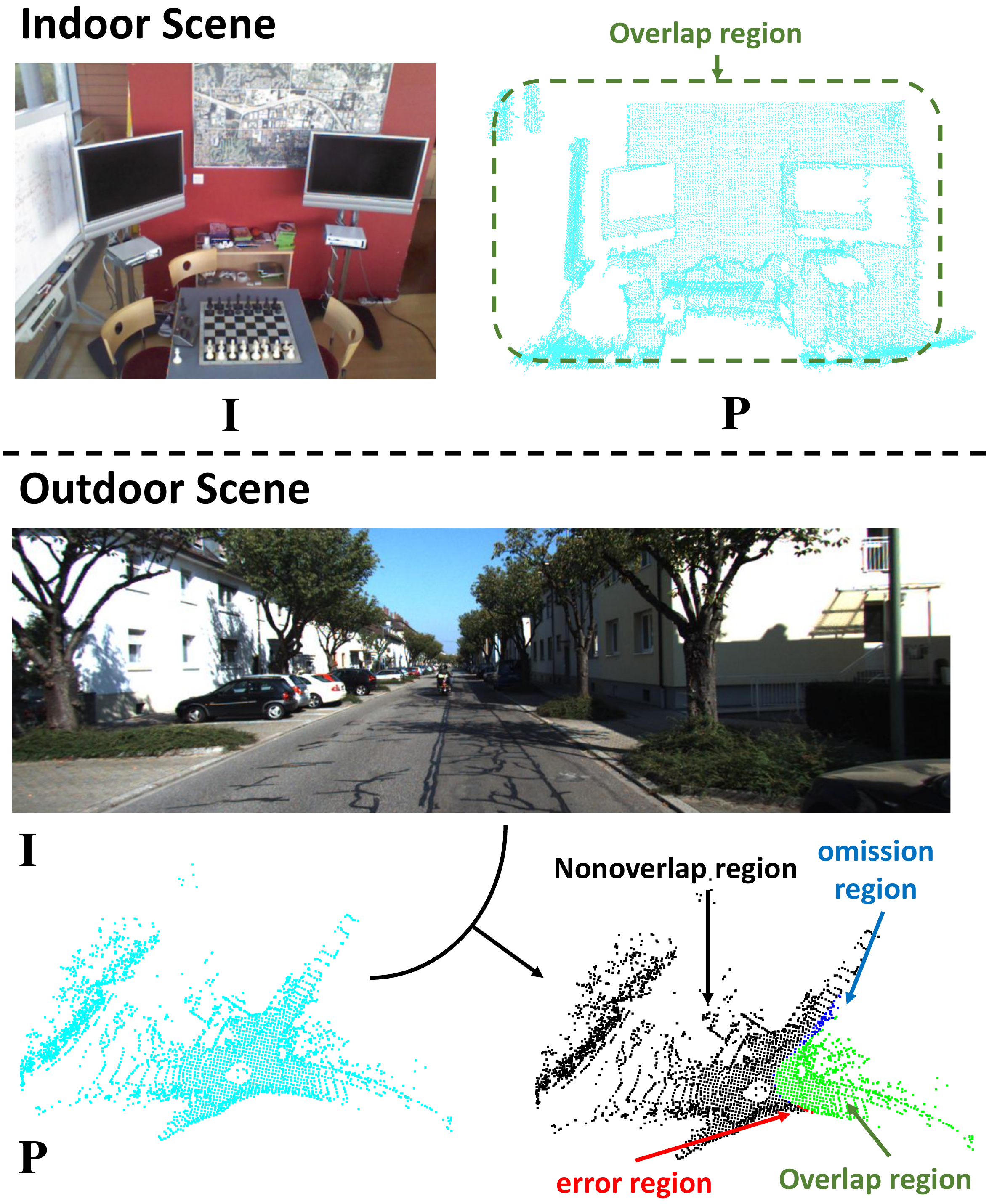}
   \vspace{-8pt}
   \caption{Indoor and outdoor scenes with images, point clouds, and overlap regions.}
   \label{fig:re}
   \vspace{-10pt}
\end{figure}

To begin with the conclusion, our approach offers a lightweight alternative to outdoor overlap region detectors, specifically addressing the small non-overlapping areas typical of indoor datasets, with minimal computational overhead and no extra detection steps.
As shown in Figure~\ref{fig:1}(a), indoor scenes (e.g., 7Scenes) exhibit substantial overlap between the point cloud and image, with few outliers. In contrast, outdoor datasets (e.g., KITTI, Figure~\ref{fig:re}(b)) have overlap limited to a small portion of the point cloud, requiring additional detectors (e.g., VP2P-match[1]). However, such detection steps inevitably introduce errors or omissions. While these are tolerable in large-scale outdoor scenes, they cause significant deviations in indoor settings. Hence, detectors designed for outdoor use not only raise computational cost indoors but also reduce registration accuracy. Indoor and outdoor methods are thus generally not directly transferable and are usually evaluated only on relevant dataset types. We believe this insight contributes meaningfully to both overall registration and fine-grained indoor alignment.

\section{Experiments}

\subsection{Datasets and Implementation Details}
Based on the 2D3D-MATR benchmark, we conducted extensive experiments and ablation studies on two challenging benchmarks: RGB-D Scenes V2 \cite{rgbdv2} and 7Scenes \cite{7scenes}.

\textbf{Dataset.} \textit{RGB-D Scenes V2} contains 14 furniture scenes. Point cloud fragments are generated from every 25 consecutive depth frames, with one RGB image sampled per 25 frames. Image-point-cloud pairs with at least 30\% overlap are selected. Scenes 1-8 are for training, 9-10 for validation, and 11-14 for testing, yielding 1,748 training, 236 validation, and 497 testing pairs.

\textit{7-Scenes} is a dataset of tracked RGB-D frames captured with a handheld Kinect at 640×480 resolution. Image-point-cloud pairs with at least 50\% overlap are selected, following the official sequence split. This results in 4,048 training, 1,011 validation, and 2,304 testing pairs.

\textbf{Implementation Details.} We used an NVIDIA Geforce RTX 3090 GPU for training. The feature dimension of the decoder output in the feature extractor is set to 512. The kernel size \( k \) is 7, and the scale value is set to the \(-0.5\) power of the number of channels. To prevent zeroing out in variance computation, a small perturbation is added. More implementation details will be provided in Appendix E.

\textbf{t-SNE.} t-Distributed Stochastic Neighbor Embedding (t-SNE) \cite{tsne} is a nonlinear dimensionality reduction technique used primarily for data visualization. It is particularly effective for visualizing high-dimensional datasets by embedding them into two or three dimensions. t-SNE aims to preserve the local structure of data points by modeling similar objects with nearby points and dissimilar objects with distant points. This method is beneficial for visualizing multimodal tasks, allowing for intuitive insights into complex datasets that span various domains such as images, text, and audio.

\textbf{MMD.} Maximum Mean Discrepancy (MMD) is a statistical method used to compare two probability distributions. It is a non-parametric technique that measures the difference between distributions by mapping data into a high-dimensional feature space using a kernel function. MMD is widely used in various machine learning applications, such as generative adversarial networks (GANs) \cite{gans}, domain adaptation, and distribution testing. The core idea of MMD is to compute the distance between the means of two distributions in a reproducing kernel Hilbert space (RKHS) \cite{rkhs}. Given two distributions \( I \) and \( Q \), the MMD is defined as:
\begin{equation}
\text{MMD}(I, Q) = \left\| \mathbb{E}_{x \sim I}[\phi(x)] - \mathbb{E}_{y \sim Q}[\phi(y)] \right\|_{\mathcal{H}},
\end{equation}

where \( \phi \) is the feature mapping function induced by a kernel, and \( \mathcal{H} \) is the RKHS.

MMD is applied in various areas, including generative models for evaluating the similarity between the distributions of generated and real data, domain adaptation for reducing distribution shifts between source and target domains, and hypothesis testing for determining if two samples are drawn from the same distribution.

\textbf{Metrics.} We evaluate our method using several key metrics:  Inlier Ratio (IR), Feature Matching Recall (FMR), and Registration Recall (RR).

\textit{Inlier Ratio} (IR) quantifies the proportion of inliers among all putative pixel-point correspondences. A correspondence is deemed an inlier if its 3D distance is less than a threshold \(\tau_1 = 5 \text{ cm}\) under the ground-truth transformation \(\mathbf{T}^*_{\mathcal{P} \rightarrow \mathcal{I}}\):
\begin{equation}
\begin{aligned}
    \text{IR} = \frac{1}{|C|} \sum_{(x_i, y_i) \in C} 
    \Big[ & \big\| \mathbf{T}^*_{\mathcal{P} \rightarrow \mathcal{I}}(x_i) \\
    & - \mathbf{K}^{-1}(y_i) \big\|_2 < \tau_1 \Big],
\end{aligned}
\end{equation}

Here, \([ \cdot ]\) denotes the Iverson bracket, \(x_i \in \mathcal{P}\), and \(y_i \in \mathcal{Q} \subseteq \mathcal{I}\) are pixel coordinates. The function \(\mathbf{K}^{-1}\) projects a pixel to a 3D point based on its depth value.

\textit{Feature Matching Recall} (FMR) represents the fraction of image-point-cloud pairs with an IR above a threshold \(\tau_2 = 0.1\). It measures the likelihood of successful registration:
\begin{equation}
\text{FMR} = \frac{1}{M} \sum_{i=1}^{M} \left[ \text{IR}_i > \tau_2 \right],
\end{equation}

where \(M\) is the total number of image-point-cloud pairs.

\textit{Registration Recall} (RR) measures the fraction of image-point-cloud pairs that are correctly registered. A pair is correctly registered if the root mean square error (RMSE) between the ground-truth-transformed and predicted point clouds \(\mathbf{T}_{\mathcal{P} \rightarrow \mathcal{I}}\) is less than \(\tau_3 = 0.1\text{ m}\):
\begin{equation}
\text{RMSE} = \sqrt{\frac{1}{|\mathcal{P}|} \sum_{p_i \in \mathcal{P}} \left\| \mathbf{T}_{\mathcal{P} \rightarrow \mathcal{I}}(p_i) - \mathbf{T}^*_{\mathcal{P} \rightarrow \mathcal{I}}(p_i) \right\|_2^2},
\end{equation}
\begin{equation}
\text{RR} = \frac{1}{M} \sum_{i=1}^{M} \left[ \text{RMSE}_i < \tau_3 \right],
\end{equation}

These metrics provide a comprehensive evaluation of the model's capability to accurately match features and register image-point-cloud pairs, ensuring robust alignment and effective correspondence.

\begin{figure}[t]
    \centering
    \includegraphics[width=\linewidth]{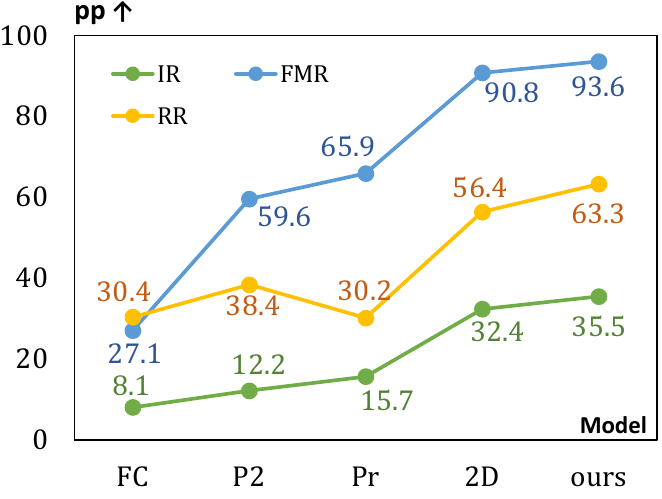}
    \caption{ Visualization of performance comparison of models on RGB-D Scenes V2. Models are abbreviated using the first two letters.}
    \label{fig:model}
\end{figure}

\begin{table}[b]
\centering
\setlength{\tabcolsep}{1mm} 
\resizebox{\columnwidth}{!}{ 
\begin{tabular}{@{}l@{\hskip 1pt}|ccccc@{}}
\toprule
\multicolumn{1}{l|}{Model} & Scene.11 & Scene.12 & Scene.13 & Scene.14 & Mean \\ \midrule
\multicolumn{1}{l|}{Mean depth (m)} & 1.74 & 1.66 & 1.18 & 1.39 & 1.49 \\ \midrule
\multicolumn{6}{c}{\textit{Inlier Ratio} ↑} \\ \midrule
\multicolumn{1}{l|}{FCGF-2D3D\cite{fcgf2d3d}} & 6.8 & 8.5 & 11.8 & 5.4 & 8.1 \\
\multicolumn{1}{l|}{P2-Net\cite{p2}} & 9.7 & 12.8 & 17.0 & 9.3 & 12.2 \\
\multicolumn{1}{l|}{Predator-2D3D\cite{predator2d3d}} & 17.7 & 19.4 & 17.2 & 8.4 & 15.7 \\
\multicolumn{1}{l|}{2D3D-MATR\cite{matr2d3d}} & \underline{32.8} & \underline{34.4} & \textbf{39.2} & \underline{23.3} & \underline{32.4} \\
\multicolumn{1}{l|}{CA-I2P (ours)} & \textbf{38.6} & \textbf{40.6} & \underline{38.9} & \textbf{24.0} & \textbf{35.5} \\ \midrule
\multicolumn{6}{c}{\textit{Feature Matching Recall} ↑} \\ \midrule
\multicolumn{1}{l|}{FCGF-2D3D\cite{fcgf2d3d}} & 11.1 & 30.4 & 51.5 & 15.5 & 27.1 \\
\multicolumn{1}{l|}{P2-Net\cite{p2}} & 48.6 & 65.7 & 82.5 & 41.6 & 59.6 \\
\multicolumn{1}{l|}{Predator-2D3D\cite{predator2d3d}} & 86.1 & 89.2 & 63.9 & 24.3 & 65.9 \\
\multicolumn{1}{l|}{2D3D-MATR\cite{matr2d3d}} & \underline{98.6} & \underline{98.0} & \underline{88.7} & \underline{77.9} & \underline{90.8} \\
\multicolumn{1}{l|}{CA-I2P (ours)} & \textbf{100.0} & \textbf{100.0} & \textbf{91.8} & \textbf{82.7} & \textbf{93.6} \\ \midrule
\multicolumn{6}{c}{\textit{Registration Recall} ↑} \\ \midrule
\multicolumn{1}{l|}{FCGF-2D3D\cite{fcgf2d3d}} & 26.5 & 41.2 & 37.1 & 16.8 & 30.4 \\
\multicolumn{1}{l|}{P2-Net\cite{p2}} & 40.3 & 40.2 & 41.2 & 31.9 & 38.4 \\
\multicolumn{1}{l|}{Predator-2D3D\cite{predator2d3d}} & 44.4 & 41.2 & 21.6 & 13.7 & 30.2 \\
\multicolumn{1}{l|}{2D3D-MATR\cite{matr2d3d}} & \underline{63.9} & \underline{53.9} & \underline{58.8} & \textbf{49.1} & \underline{56.4} \\
\multicolumn{1}{l|}{CA-I2P (ours)} & \textbf{68.1} & \textbf{73.5} & \textbf{63.9} & \underline{47.8} & \textbf{63.3} \\ \bottomrule
\end{tabular}
}
\vspace{-8pt}
\caption{Evaluation results on RGB-D Scenes V2. \textbf{Boldfaced} numbers highlight the best and the second best are \underline{underlined}.}
\label{tab:rgbdv2}
\end{table}

\begin{table}[t]
\centering
\small
\setlength{\tabcolsep}{2pt}
\begin{tabular}{lccc}
\hline
Model & FMR $\uparrow$ & IR $\uparrow$ & RR $\uparrow$ \\
\hline
FreeReg+Kabsch~\cite{freereg}  & 82.0 & 30.9 & 34.1 \\
FreeReg+PnP~\cite{freereg}     & 81.5 & 29.7 & 57.3 \\
\textbf{CA-I2P (Ours)}          & \textbf{93.6} & \textbf{35.5} & \textbf{63.3} \\
\hline
\end{tabular}
\vspace{-5pt}
\caption{More experiments on RGB-D Scenes V2.}
\label{tab1}
\vspace{-5pt}
\end{table}

\begin{table}[ht]
\centering
\setlength{\tabcolsep}{1mm} 
\begin{tabular}{lcccccccc}
\toprule
\multicolumn{1}{l|}{Mtd}  & Chs & Fr & Hds & Off & Pmp & Ktn & Strs & Mean \\ \midrule
\multicolumn{1}{l|}{Mdpt} & 1.78 & 1.55 & 0.80 & 2.03 & 2.25 & 2.13 & 1.84 & 1.77 \\ \midrule
\multicolumn{9}{c}{\textit{Inlier Ratio} \(\uparrow\)} \\ \midrule
\multicolumn{1}{l|}{FC\cite{fcgf2d3d}} & 34.2 & 32.8 & 14.8 & 26 & 23.3 & 22.5 & 6.0 & 22.8 \\
\multicolumn{1}{l|}{P2\cite{p2}} & 55.2 & 46.7 & 13.0 & 36.2 & 32.0 & 32.8 & 5.8 & 31.7 \\
\multicolumn{1}{l|}{Pr\cite{predator2d3d}} & 34.7 & 33.8 & 16.6 & 25.9 & 23.1 & 22.2 & 7.5 & 23.4 \\
\multicolumn{1}{l|}{2D\cite{matr2d3d}} & \underline{72.1} & \underline{66.0} & \underline{31.3} & \underline{60.7} & \underline{50.2} & \underline{52.5} & \underline{18.1} & \underline{50.1} \\
\multicolumn{1}{l|}{ours} & \textbf{73.6} & \textbf{66.4} & \textbf{34.5} & \textbf{62.4} & \textbf{52.1} & \textbf{52.8} & \textbf{19.1} & \textbf{51.6} \\ \midrule
\multicolumn{9}{c}{\textit{Feature Matching Recall} \(\uparrow\)} \\ \midrule
\multicolumn{1}{l|}{FC\cite{fcgf2d3d}} & \underline{99.7} & 98.2 & 69.9 & 97.1 & 83.0 & 87.7 & 16.2 & 78.8 \\
\multicolumn{1}{l|}{P2\cite{p2}} & \textbf{100.0} & 99.3 & 58.9 & \underline{99.1} & 87.2 & 92.2 & 16.2 & 79 \\
\multicolumn{1}{l|}{Pr\cite{predator2d3d}} & 91.3 & 95.1 & \underline{76.6} & 88.6 & 79.2 & 80.6 & 31.1 & 77.5 \\
\multicolumn{1}{l|}{2D\cite{matr2d3d}} & \textbf{100.0} & \underline{99.6} & \textbf{98.6} & \textbf{100.0} & \textbf{92.4} & \textbf{95.9} & \underline{58.2} & \underline{92.1} \\
\multicolumn{1}{l|}{ours} & \textbf{100.0} & \textbf{100.0} & \textbf{98.6} & \textbf{100.0} & \underline{92.0} & \underline{95.5} & \textbf{60.8} & \textbf{92.4} \\ \midrule
\multicolumn{9}{c}{\textit{Registration Recall} \(\uparrow\)} \\ \midrule
\multicolumn{1}{l|}{FC\cite{fcgf2d3d}} & 89.5 & 79.7 & 19.2 & 85.9 & 69.4 & 79.0 & 6.8 & 61.4 \\
\multicolumn{1}{l|}{P2\cite{p2}} & \underline{96.9} & \underline{86.5} & 20.5 & 91.7 & 75.3 & 85.2 & 4.1 & 65.7 \\
\multicolumn{1}{l|}{Pr\cite{predator2d3d}} & 69.6 & 60.7 & 17.8 & 62.9 & 56.2 & 62.6 & 9.5 & 48.5 \\
\multicolumn{1}{l|}{2D\cite{matr2d3d}} & \underline{96.9} & \textbf{90.7} & \underline{52.1} & \underline{95.5} & \underline{80.9} & \underline{86.1} & \underline{28.4} & \underline{75.8} \\
\multicolumn{1}{l|}{ours} & \textbf{99.0} & \textbf{90.7} & \textbf{68.5} & \textbf{96.2} & \textbf{83.0} & \textbf{88.1} & \textbf{31.1} & \textbf{79.5} \\ \bottomrule
\end{tabular}
\caption{Evaluation results on 7Scenes. Models are abbreviated using the first two letters, and scenes are abbreviated. \textbf{Boldfaced} numbers highlight the best and the second best are \underline{underlined}.}
\vspace{-8pt}
\label{tab:7}
\end{table}
 
\subsection{Evaluations on Dataset}

We compared our approach with 2D3D-MATR and other baselines on the RGB-D Scenes V2 dataset (Table~\ref{tab:rgbdv2}). Our method integrates the Channel Adaptive Adjustment Module (CAA) to optimize image and point cloud features along the channel dimension, effectively reducing the cross-modal domain gap and preventing attention to non-overlapping regions. Specifically, the CAA consists of the Intra-Modal Enhancement Stage (IME) and the Cross-Modal Channel Filtering Stage (CMCF). IME leverages rotation and residual transformations to establish inter-channel dependencies for image features, while applying self-attention mechanisms to enhance the channel dimension for point cloud features. CMCF further refines feature representations by identifying and masking incompatible channels using the covariance matrix, thereby preserving matching regions while mitigating domain discrepancies. In addition, we introduce the Global Optimal Selection (GOS) module, which replaces the traditional top-k selection method with an optimal transport-based approach. This enables more accurate global feature association, reducing multi-to-one matching errors and improving overall registration robustness. Through these enhancements, our approach achieves a 3.1 percentage point (pp) gain in inlier ratio and a 2.8 pp increase in feature matching recall. Most notably, our method surpasses the previous state-of-the-art in registration recall by 6.9 pp, demonstrating its effectiveness in bridging the gap between 2D image and 3D point cloud data. We compare with recent baselines (Table~\ref{tab1}), and our method outperforms FreeReg by 6 pp  in RR.

Compared to RGB-D Scenes V2, 7-Scenes exhibit greater scale variations, yet our method still outperforms others, as shown in Table~\ref{tab:7}. We achieve a 3.7 percentage point advantage over 2D3D-MATR in Registration Recall, reaching 79.5\%. Notably, CA-I2P improves performance in the challenging Heads and Stairs scenes. In the Heads scene, small 3D errors are amplified due to close camera proximity, while the Stairs scene contains repetitive patterns that complicate matching. CA-I2P effectively addresses these challenges through enhanced feature representation and context-aware fusion, ensuring better alignment and robustness across diverse indoor environments.

\subsection{Ablation Studies}

We conducted ablation studies on the RGB-D Scenes V2 dataset to evaluate the effectiveness of CAA (comprising ICE and PCE) and GOS, with the results presented in Table~\ref{tab:ablation}. To further analyze the impact of similarity filtering, we introduced a new metric, PIR, which quantifies the proportion of correctly matched points among all correspondences, providing a more direct measure of the quality and reliability of I2P registration. Additional ablation studies can be found in Appendix I.

\begin{table}[t]
\centering
\resizebox{\columnwidth}{!}{
\begin{tabular}{c||cccc||cccc}
\toprule
Method & ICE & PCE & CMCF & GOS & PIR & IR & FMR & RR \\
\midrule
M1 &   &   &   &   & 48.5 & 32.5 & 91.0 & 55.8 \\
M2 & \checkmark &   &   &  & 56.3 & 34.6 & 92.4 & 56.9 \\
M3 &   & \checkmark &   &   & 54.7 & 33.6 & 93.2 & 56.0 \\
M4 & \checkmark & \checkmark  &   &   & 56.3 & 34.6 & 92.3 & 58.2 \\
M5 &   &   & \checkmark &   & \colorbox{yellow}{\textbf{59.2}} & 35.4 & 91.4 & 59.7 \\
M6 & \checkmark  & \checkmark  & \checkmark &   & 59.1 & 35.1 & 93.3 & 61.8 \\
M7 &   &   &   & \checkmark & 56.3 & 35.3 & 91.7 & 58.1 \\
M8 & \checkmark & \checkmark & \checkmark & \checkmark & 58.3 & \colorbox{yellow}{\textbf{35.5}} & \colorbox{yellow}{\textbf{93.6}} & \colorbox{yellow}{\textbf{63.3}} \\
\bottomrule
\end{tabular}}
\vspace{-10pt}
\caption{Ablation study results. CAA includes ICE, PCE, and CMCF , GOS indicates our method for selection. PIR, IR, FMR, and RR denote evaluation metrics.}
\label{tab:ablation}
\end{table}

First, we analyzed intra-modal feature enhancement: M2, M3, and M4, representing image, point cloud, and joint feature enhancement, respectively, all showed improvements, with M4 increasing RR by 2.3pp.In M5 and M6, cross-modal channel filtering improves matching, with M5 achieving a 10.7pp PIR boost. In M7, GOS reduces many-to-one matching, increasing IR by 2.8pp. The final version, M8, enhances RR by 7.5pp.

\begin{table}[ht]
\centering
\setlength{\tabcolsep}{3pt} 
\renewcommand{\arraystretch}{1.1} 
\resizebox{\columnwidth}{!}{ 
\begin{tabular}{lc|cccc}
\toprule
\multicolumn{2}{l|}{ICE}                                               & PIR           & FMR           & IR           & RR           \\ \midrule

\multicolumn{2}{l|}{2D3D-MATR (baseline)}                              & 48.5          & 91.0          & 32.5          & 55.8          \\ \midrule
\multicolumn{2}{l|}{w/ \( F_{I1} \)}              & 51.4          & 92.2          & 33.3          & 55.5          \\ \midrule
\multicolumn{2}{l|}{w/ \( F_{I1} \) \& \( F_{I2} \)} & 52.2          & 91.7          & 34.5          & 56.0          \\ \midrule
\multicolumn{2}{l|}{w/ \( F_{I1} \) \& \( F_{I3} \)} & 55.2          & 92.0          & 34.4          & 56.5          \\ \midrule
\multicolumn{2}{l|}{w/ \( F_{I2} \) \& \( F_{I3} \)} & 54.2          & 91.8          & 33.9          & 55.9          \\ \midrule   
\multicolumn{2}{l|}{w/ \( F_{I1} \) \& \( F_{I2} \) \& \( F_{I3} \)}   
                & \colorbox{yellow}{\textbf{56.3}} & \colorbox{yellow}{\textbf{92.4}} & \colorbox{yellow}{\textbf{34.6}} & \colorbox{yellow}{\textbf{56.9}} \\ \bottomrule
\end{tabular}
}
\caption{Ablation studies of Image Channel Enhancement Unit (ICE) on RGB-D Scenes V2. CAA includes ICE, PCE, and CMCF, while GOS denotes our method for selection. PIR, IR, FMR, and RR represent evaluation metrics.}
\label{tab:performance}
\end{table}

Next, we conducted an ablation study on the RGB-D Scenes V2 dataset to evaluate the impact of the Image Channel Enhancement Unit (ICE) on matching performance, as shown in Table~\ref{tab:performance}. Starting with the baseline model 2D3D-MATR, the PIR, FMR, IR, and RR scores are 48.5, 91.0, 32.5, and 55.8, respectively. When introducing only $F_{I1}$ for channel enhancement, PIR and IR slightly improved to 51.4 and 32.3, while FMR increased to 92.2, indicating that the enhanced features improved feature matching recall. Further incorporating $F_{I2}$ resulted in a PIR increase to 52.2 and an RR improvement to 56.0, demonstrating that integrating cross-channel information benefits the accuracy of image-to-point cloud registration. When adding $F_{I3}$ on top of $F_{I1}$, PIR and IR continued to rise to 55.2 and 34.4, while FMR remained stable at 92.0, suggesting that this enhancement strategy effectively improves matching quality. Similarly, the combination of $F_{I2}$ and $F_{I3}$ provided additional gains, bringing RR up to 55.9. However, the optimal configuration was achieved when using $F_{I1}$, $F_{I2}$, and $F_{I3}$ together, leading to a PIR of 56.3, IR of 34.6, and RR of 56.9. Compared to the baseline, this configuration resulted in a significant overall improvement, demonstrating that joint channel-level enhancement maximizes the stability and accuracy of image-to-point cloud matching.

These results indicate that ICE effectively enhances feature robustness by integrating multiple channel enhancement strategies, reducing mismatches and improving overall registration accuracy. When further combined with PCE, CMCF, and GOS for global optimization, registration performance can be further improved in terms of stability and robustness.

\begin{table}[ht]
\centering
\setlength{\tabcolsep}{8pt} 
\renewcommand{\arraystretch}{1.2} 
\begin{tabular}{l|c}
\hline
Method  & Num of Parameters \\ \hline
Base    & 31.05M  \\ 
ICE     & 31.06M  \\ 
PCE     & 31.24M  \\ 
ICE \& PCE & 31.25M  \\ \hline
\end{tabular}
\caption{Comparison of the number of parameters for Intra-Modal Enhancement stage.}
\label{tab:params}
\end{table}

Table~\ref{tab:params} compares the number of parameters across different configurations of our model. The results indicate that the introduction of ICE and PCE leads to only a marginal increase in the total number of parameters, with the largest growth being just 0.2M (from 31.05M to 31.25M). This minimal change suggests that our proposed method does not rely on significantly increasing model complexity to achieve performance improvements. Instead, the enhanced registration accuracy stems from the effective refinement of feature representations and the improved matching strategy introduced by ICE and PCE. By leveraging channel-adaptive adjustments and global optimal selection, our approach optimizes feature alignment without adding substantial computational overhead, demonstrating its efficiency and practicality in multimodal registration tasks.

As shown in Table~\ref{tab2}, we report inference time to reflect complexity. 
Although the overall inference time increases due to the added computational complexity, the matching time is reduced as the proposed OT-based filtering effectively eliminates low-quality correspondences early, reducing redundant computations. Considering the performance gain, the additional cost is acceptable. 
\begin{table}[ht]
\centering
\small
\setlength{\tabcolsep}{2pt}
\begin{tabular}{lcc}
\hline
Module & Matching Time (s) & Total Inference Time (s)\\
\hline
w/o GOS & 0.124 & 1.271\\
w GOS   & 0.080 & 2.837 \\
\hline
\end{tabular}
\vspace{-5pt}
\caption{Comparison of inference time.}
\label{tab2}
\vspace{-15pt}
\end{table}

We also conducted extensive ablation studies on the 7Scenes dataset to further evaluate the effectiveness of each module, as shown in Table~\ref{tab:ablation_7scenes}. The ablation study investigates the contributions of the Intra-Modal Enhancement Stage (IME), Cross-Modal Channel Filtering Stage (CMCF), and Global Optimal Selection (GOS) in improving image-to-point cloud registration.

\begin{table}[b]
\centering
\setlength{\tabcolsep}{1mm} 
\resizebox{\columnwidth}{!}{ 
\begin{tabular}{clc|c|c||cccc}
\toprule
\multicolumn{2}{c|}{Model} & IME & CMCF & GOS & PIR  &  FMR  & IR  & RR   \\ 
\midrule
\multicolumn{2}{c|}{F1} &  &  &  & 83.9 & 92.1 & 50.1 & 75.8 \\ 
\midrule
\multicolumn{2}{c|}{F2} & \checkmark &  &  & 83.4 & \textbf{93.3} & 51.4 & 77.3 \\ 
\midrule
\multicolumn{2}{c|}{F3} & \checkmark & \checkmark &  & \textbf{84.3} & 92.3 & 51.5 & 79.2 \\ 
\midrule
\multicolumn{2}{c|}{F4} &  &  & \checkmark & 83.5 & 91.7 & 50.5 & 76.9 \\ 
\midrule
\multicolumn{2}{c|}{F5} & \checkmark & \checkmark & \checkmark & 83.9 & 92.4 & \textbf{51.6} & \textbf{79.5} \\ 
\bottomrule
\end{tabular}
}
\caption{Ablation studies on 7Scenes.}
\label{tab:ablation_7scenes}
\end{table}
\begin{figure*}[ht]
    \centering
    \includegraphics[width=\linewidth]{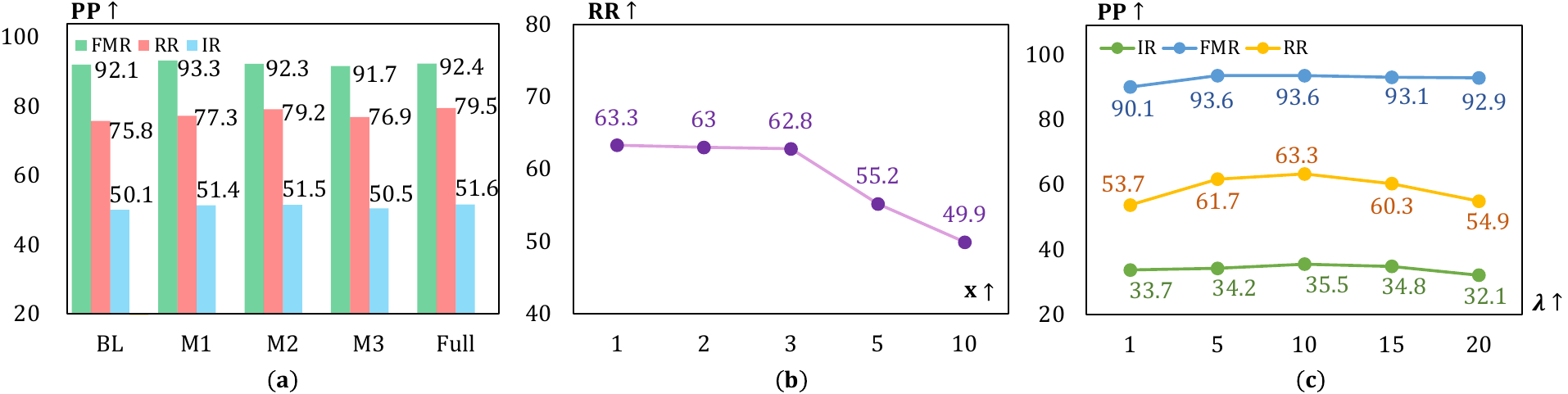} 
    \caption{Visualization of ablation studies.}
    \label{fig:multi}
\end{figure*}

Starting with the baseline model F1, which does not incorporate any additional enhancement modules, the PIR, FMR, IR, and RR scores are 83.9, 92.1, 50.1, and 75.8, respectively. When introducing IME in F2, feature refinement within the same modality improves, leading to a slight increase in PIR to 83.4 and a notable rise in FMR to 93.3. This demonstrates that intra-modal enhancement effectively strengthens feature representations, making feature matching more reliable.
Further incorporating CMCF in F3 refines feature selection by filtering out incompatible channels, resulting in a PIR increase to 84.3 and an IR improvement to 51.5. The rise in IR indicates that cross-modal feature filtering enhances the identification of correct correspondences between images and point clouds, reducing mismatches. However, FMR slightly decreases to 92.3, suggesting that filtering redundant channels slightly reduces recall but improves precision.
In F4, the addition of GOS replaces the conventional top-k selection with optimal transport, allowing for more precise global matching. Although PIR remains relatively stable at 83.5, the IR improvement to 50.5 and RR increase to 76.9 demonstrate that reducing many-to-one matching errors contributes to better overall registration performance.
The final model, F5, integrates IME, CMCF, and GOS, achieving the highest performance across all metrics. Compared to the baseline, F5 reaches a PIR of 83.9, FMR of 92.4, IR of 51.6, and RR of 79.5. The 1.5 percentage point gain in RR over the baseline highlights the cumulative effectiveness of all three modules in improving robustness and accuracy in multimodal registration. We present the visualization results in Figure~\ref{fig:multi} (a).

These results confirm that IME, CMCF, and GOS each play a crucial role in enhancing image-to-point cloud matching. IME strengthens intra-modal feature quality, CMCF filters out noisy and redundant features, and GOS optimizes global matching. When combined, these components significantly improve registration recall, ensuring more accurate and robust correspondences in challenging indoor environments.

We also conducted experiments to explore the impact of the number of CMCF iterations \( x \) and the Sinkhorn iterations \( \lambda \) on RGB-D Scenes V2. These studies aim to understand the impact of these parameters on model performance and to optimize their settings for better results.

Figure \ref{fig:multi} (b) illustrates the impact of the number of CMCF iterations x values on the model's performance, measured by RR, within the RGB-D Scenes V2 dataset. The X-axis represents different values of x, while the Y-axis indicates the corresponding RR values. It is observed that the model achieves the highest registration recall (RR) of 63.3 when the number of CMCF iterations is set to 1. As the number of iterations increases, the RR value gradually decreases, reaching 49.9 at 10 iterations. This suggests that while a small number of iterations enhances registration performance, excessive iterations may introduce unnecessary filtering, leading to a decline in accuracy. Consequently, a lower iteration count is preferable for maintaining optimal RR. Additionally, the observed variation in RR across different iteration settings indicates the model’s sensitivity to excessive filtering while demonstrating its ability to maintain reasonable performance within a controlled range of iterations.

Figure \ref{fig:multi} (c) depicts the effect of varying the sinkhorn iterations $\lambda$ on key performance metrics (FMR, IR, RR) within the RGB-D Scenes V2 dataset. The X-axis shows different values of $\lambda$, while the Y-axis corresponds to the associated performance metrics. As $\lambda$ increases from 1 to 20, FMR remains relatively stable, fluctuating between 90.1 and 93.6, indicating that feature matching recall is minimally affected by changes in $\lambda$. In contrast, RR exhibits a noticeable peak at $\lambda=10$, reaching 63.3, followed by a gradual decline as $\lambda$ increases further, suggesting that an optimal balance is achieved at this setting. Similarly, IR also experiences a slight increase up to $\lambda=10$, peaking at 35.5, before decreasing to 32.1 at $\lambda=20$. 

These results suggest that a moderate $\lambda$ value, particularly around 10, leads to the best overall registration performance, maximizing RR and IR while maintaining a high FMR. Setting $\lambda$ too low may limit the effectiveness of feature selection, whereas excessively high values could introduce unnecessary constraints, reducing the model’s ability to generalize across different image-point cloud pairs.

\begin{figure}[b]
    \centering
    \includegraphics[width=1\linewidth]{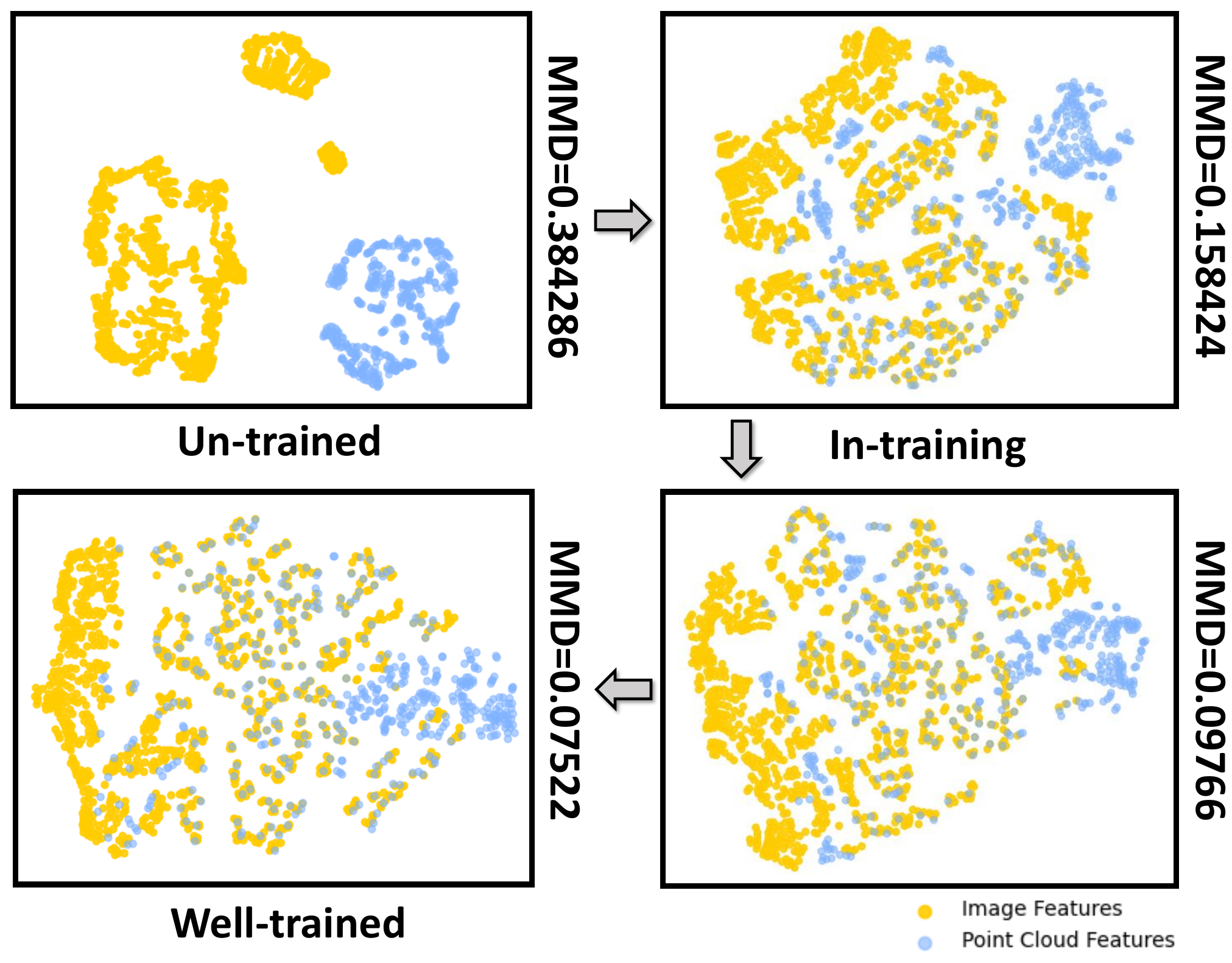} 
    \caption{Visualization of modality differences. With training, the modalities and distributions of the point cloud and image become more aligned, reducing the domain gap.}
    \label{fig:4}
\end{figure}

\subsection{Visualization}

We use visualization to intuitively demonstrate CA-I2P’s performance. As shown in Figure \ref{fig:4}, the four t-SNE plots depict image and point cloud features at different training stages (untrained, training, and trained), highlighting modal differences. The maximum mean discrepancy (MMD) at each stage reflects domain shifts caused by CAA. Initially, image features are dense while point cloud features are sparse, but training reduces the modal gap for a more uniform distribution. The channel enhancement filtering module ensures feature consistency, enhancing registration efficiency, robustness, and generalization.

\begin{figure}[ht]
    \centering
    \begin{subfigure}{\columnwidth}
        \centering
        \includegraphics[width=\linewidth]{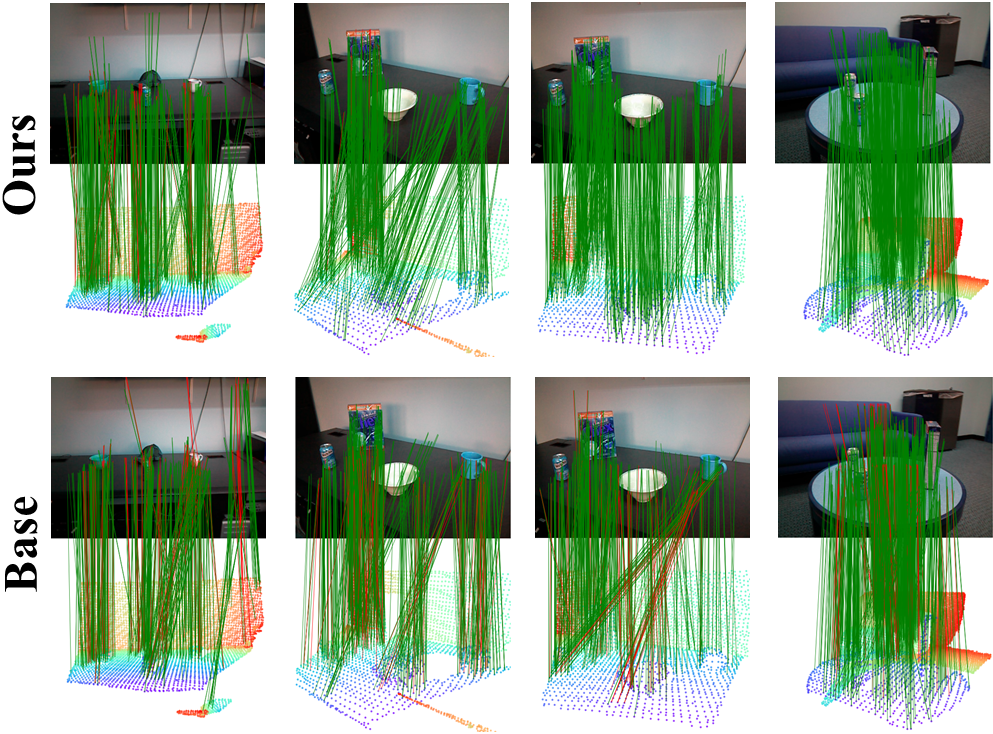}
        \caption{Visualization on RGB-D Scenes V2}
        \label{fig:r_visual}
    \end{subfigure}
    
    \vspace{2pt} 
    
    \begin{subfigure}{\columnwidth}
        \centering
        \includegraphics[width=\linewidth]{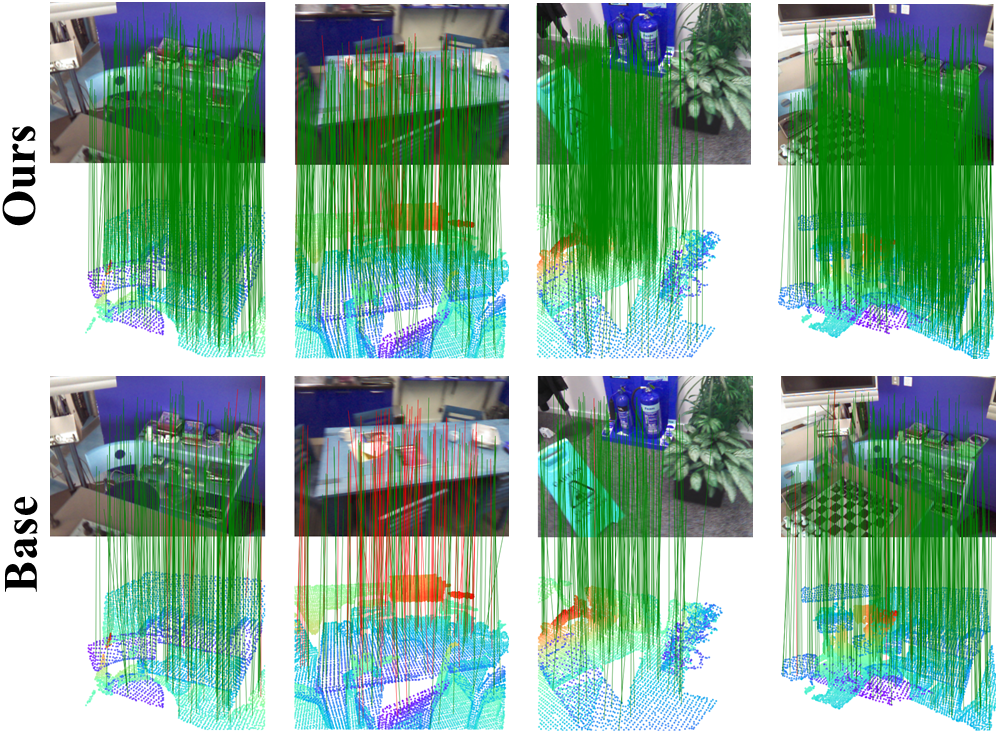}
        \caption{Visualization on 7Scenes}
        \label{fig:7_visual}
    \end{subfigure}
    \caption{The visualization result of CA-I2P}
    \label{fig:5}
            \vspace{-0.4cm}
\end{figure}

Figure \ref{fig:5} visualizes the results of CA-I2P, where lines indicate correspondences—green for matches within the threshold and red for those exceeding it. After feature optimization, our method significantly enhances matching performance by establishing a greater number of accurate correspondences while reducing erroneous matches. This improvement is reflected in the increased number of green lines and the noticeable reduction of red lines, indicating higher matching accuracy. Furthermore, the reduction of redundant matches contributes to a more robust and reliable registration process. Additional visualizations and qualitative results can be found in Appendix J.

\begin{figure}[ht]
  \centering
  \small
  \includegraphics[width=0.8\linewidth]{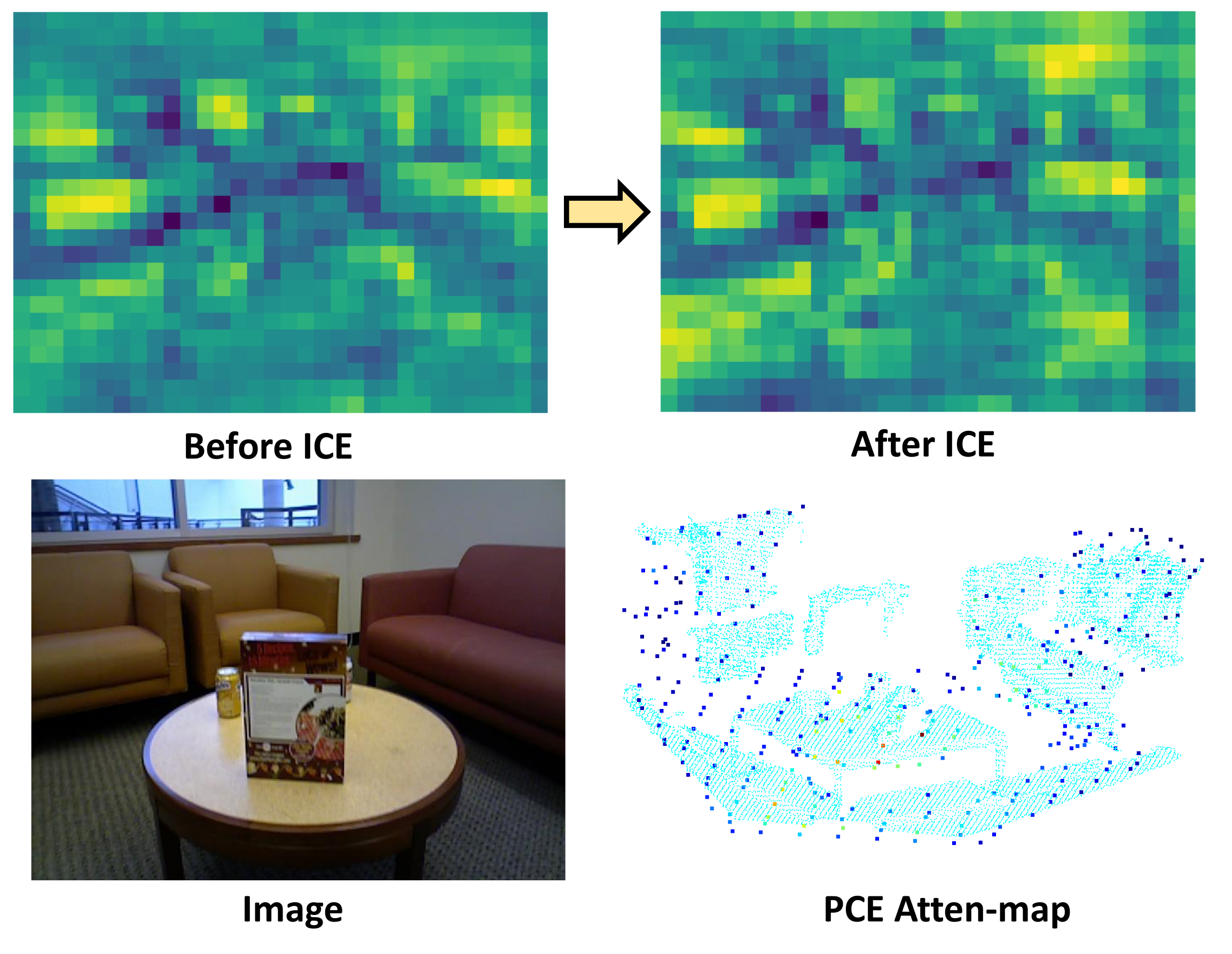}
   \vspace{-10pt}
   \caption{Visualzation of intra-model enhancement stage.}
    \vspace{-10pt}
   \label{fig:re3}
\end{figure}

As shown in Figure~\ref{fig:re3}, it can be observed that after the ICE unit, the image features in previously overlooked regions are enhanced, leading to an overall improvement in feature quality (e.g., overall brightness increases). For the PCE module, the visualized attention maps demonstrate that important regions for registration, such as the table and sofa, are more prominently attended to by the attention mechanism.
\section{Conclusion}
We introduce CA-I2P, an innovative image-point cloud registration network designed to optimize feature representation in the channel dimension with globally optimal selection. Our method enhances intra-modal feature channels and filters cross-modal feature channels, improving the representation of overlapping regions in different modal features. Furthermore, we implement a globally optimal selection process to minimize potential redundant matches. CA-I2P demonstrates state-of-the-art performance on the RGB-D Scenes V2 and 7-Scenes datasets.

{
    \small
    \bibliographystyle{ieeenat_fullname}
    \bibliography{main}
}

\end{document}